\documentclass[conference]{IEEEtran}
\IEEEoverridecommandlockouts
\IEEEpubid{\makebox[\columnwidth]{978-1-7281-6232-4/20/\$31.00~\copyright2020 IEEE \hfill} \hspace{\columnsep}\makebox[\columnwidth]{ }}

\usepackage{cite}
\usepackage{amsmath,amssymb,amsfonts}
\usepackage{algorithmic}
\usepackage{graphicx}
\usepackage{textcomp}
\usepackage{xcolor}
\def\BibTeX{{\rm B\kern-.05em{\sc i\kern-.025em b}\kern-.08em
    T\kern-.1667em\lower.7ex\hbox{E}\kern-.125emX}}
\begin{document}

\title{Forensic Authorship Analysis of Microblogging Texts Using $N$-Grams and Stylometric Features
\thanks{This work has been carried out by Nicole Belvisi in the context of their Bachelor Thesis at Halmstad University.}
}

\author{\IEEEauthorblockN{Nicole Mariah Sharon Belvisi}
\IEEEauthorblockA{\textit{School of Information Technology (ITE)} \\
\textit{Halmstad University}, Sweden\\
nicbel18@student.hh.se}
\and
\IEEEauthorblockN{Naveed Muhammad}
\IEEEauthorblockA{\textit{Institute of Computer Science} \\
\textit{University of Tartu, Estonia}\\
naveed.muhammad@ut.ee}
\and
\IEEEauthorblockN{Fernando Alonso-Fernandez}
\IEEEauthorblockA{\textit{School of Information Technology (ITE)} \\
\textit{Halmstad University}, Sweden\\
feralo@hh.se}
}

\maketitle

\begin{abstract}
In recent years, messages and text posted on the Internet are used in criminal investigations. Unfortunately, the authorship of many of them remains unknown.
In some channels, the problem of establishing authorship may be even harder, since the length of digital texts is limited to a certain number of characters.
In this work, we aim at identifying authors of tweet messages, which are limited to 280 characters. We evaluate popular features employed traditionally in authorship attribution which capture properties of the writing style at different levels. We use for our experiments a self-captured database of 40 users, with 120 to 200 tweets per user. Results using this small set are promising, with the different features providing a classification accuracy between 92\% and 98.5\%. These results are competitive in comparison to existing studies which employ short texts such as tweets or SMS.
\end{abstract}


\section{Introduction}

New technologies have brought new ways of communication.
Internet, social media, SMS, emails and other applications have enabled faster and more efficient way to deliver messages.
More importantly, they have also given the gift of anonymity.
Unfortunately, cyber-criminals can also take advantage of such anonymity for malicious purposes.
In this context, authorship analysis becomes an important issue in forensic investigation,
finding application in tasks aimed to counteract cyberbullying, cyberstalking, fraud, or ransom notes, to name just a few \cite{Nirkhi2013}.

In the early years before the ``Tech Era'', messaging was primarily done in handwriting forms. However, it is not possible to use handwriting analysis with digital communication methods to look for authorship evidence. Moreover, some channels limit the length of messages to just a few hundred characters, making the problem even more challenging \cite{Donais13,Ragel13,Okuno14}.
An added difficulty is that nowadays, new elements such as slang words, shortcuts or emojis are common, and they change over time according to trends.
In order to determine the identity of an individual, the analyst would often resort to geo-location or IP addresses. Nonetheless, it is possible to conceal such elements easily. In such cases, the text is the only evidence available, making necessary to develop new tools and techniques to determine authorship from such pieces of digital evidence.
Accordingly, the purpose of this work is to analyze methods to identify the writer of a digital text. We will focus on short texts limited to 280 characters (Twitter posts).

Authorship analysis aims at establishing a connection between a text and its author. It relies on the fact that every person has a specific set of features in their writing style that distinguishes them from any other individual \cite{[Srihari02]}.
The problem can be approached from three different perspectives \cite{Nirkhi2013,Bouanani2014}:

\begin{itemize}
  \item \textit{Authorship Identification}, also known as \textit{Authorship Attribution}. This is concerned with finding the author of an anonymous text. Features and patterns of the anonymous text are compared against a database of texts whose authors are known. The closest author of the database are assigned as the author of the unknown text.  

  \item \textit{Authorship Verification}. 
  Given two pieces of text, one anonymous and one written by a known author, the tasks tries to answer if the two texts have been written by the same person. 
  This finds application e.g. in plagiarism detection, or in different persons claiming to be the author of a text \cite{HOLMES95}.

  \item \textit{Profiling}. This task aims at constructing a psychological/sociological profile of the author.
  A text can reveal not only the identity of its writer, but it is believed that it can provide other 
  information such as the level of education, the age, or country of origin \cite{Bouanani2014}.

\end{itemize}

\begin{figure}[htb]
\centering
        \includegraphics[width=0.36\textwidth]{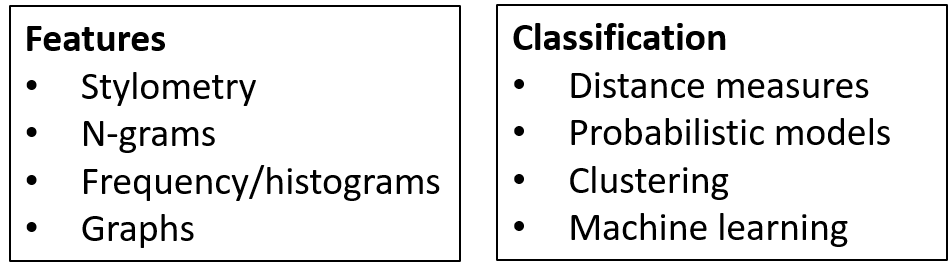}
\caption{Summary of features and classification methods employed for authorship identification in the literature.}
\label{fig:SOA}
\end{figure}

\begin{table*}[htb]
\scriptsize
\begin{center}
\begin{tabular}{cccccc}

\multicolumn{6}{c}{} \\

Ref & Year & Focus & Data & Features & Classification  \\ \hline 

\cite{Abbasi05} & 2005  & Forums (Arabic)  & 20 persons, 20 messages/each &  Stylometry & DTs, SVMs \\ \hline

\cite{Mohtasseb09} & 2009  & Blogs & 93 persons, 17647 post, 200 posts/person  & Stylometry &  SVM, NB \\ \hline

\cite{Pillay10} & 2010 & Forums & 5-100 persons, 17-161 messages/each & Stylometry & Clustering, DT, NB \\ \hline

\cite{Koppel11} & 2011  &  Blogs & 10000 blogs, 2000 words/each  & $n$-grams & Cosine distance \\ \hline

\cite{Cristani12} & 2012  & Chat (Italian)  & 77 persons, 615 words/each  &  Stylometry & Bhattacharya/Euclidean distance \\ \hline

\cite{Amuchi12} & 2012 & Chat & 10 persons, 341 posts in total  & Stylometry, $n$-grams & Chi-Square distance, NB, SVM \\ \hline

\cite{Donais13} & 2013 & SMS  & 81 persons, 2000 SMS, $<$50 SMS/person  & n/a &  Modified NB \\ \hline

\cite{Ragel13} & 2013  & SMS (Eng, Ch)  & 70 persons, $>$50 SMS/person   & $n$-grams & Cosine/Euclidean distance \\  \hline

\cite{Inches13} & 2013 & Chat & 4.6k-19k persons, 79k-93k messages in total & Term frequency &  Chi-Square dist., KL divergence \\ \hline

\cite{Marukatat14} & 2014 & Forums (Thai) & 6 persons, 25 messages/person, 143 words/message  & Stylometry &  SVM, DT \\ \hline

\cite{Johnson14} & 2014  & Emails & 176 persons, 63000 emails in total  &  Stylometry, $n$-grams & Jaccard coefficient \\ \hline

\cite{Okuno14} & 2014  & Tweets & 10000 persons, 120 tweets/person  & $n$-grams  & Cosine distance \\ \hline

\cite{Yang14} & 2014 & Blogs & 19320 persons, 678161 post, 7250 words/person & Stylometry, $n$-grams &  LR \\ \hline

\cite{Nirmal15} & 2015 & Emails & 50 persons, $>$200 emails/person & Graphs, Node frequency, Probability Models  &  SVM   \\ \hline


\end{tabular}

\end{center}
\caption{Existing studies in online authorship identification. DT= Decission Trees. SVM=Support Vector Machines. LR= Logistic Regression. NB=Naive Bayes. NN=Neural Networks. KL=Kullback-Leibler Eng=English. Ch=Chinese Mandarin.}
\label{tab:SOA}
\end{table*}

\normalsize

In this work, we will focus on authorship identification or attribution.
It is one of the major and most influential issues in digital forensics,
given the opportunity that today’s technology has enabled
to freely communicate without revealing our identity \cite{Nirkhi2013}.
Authorship analysis of texts has its origin in a linguistic research area called stylometry, which refers to statistical analysis of literary style \cite{Williams75,Nirkhi2013}.
Today, an increasing amount of research is focused on the
analysis of online messages
due to the growth of web applications and social networks \cite{Nirkhi2013,Bouanani2014,Stamatatos2009}.

\begin{figure}[htb]
\centering
        \includegraphics[width=0.4\textwidth]{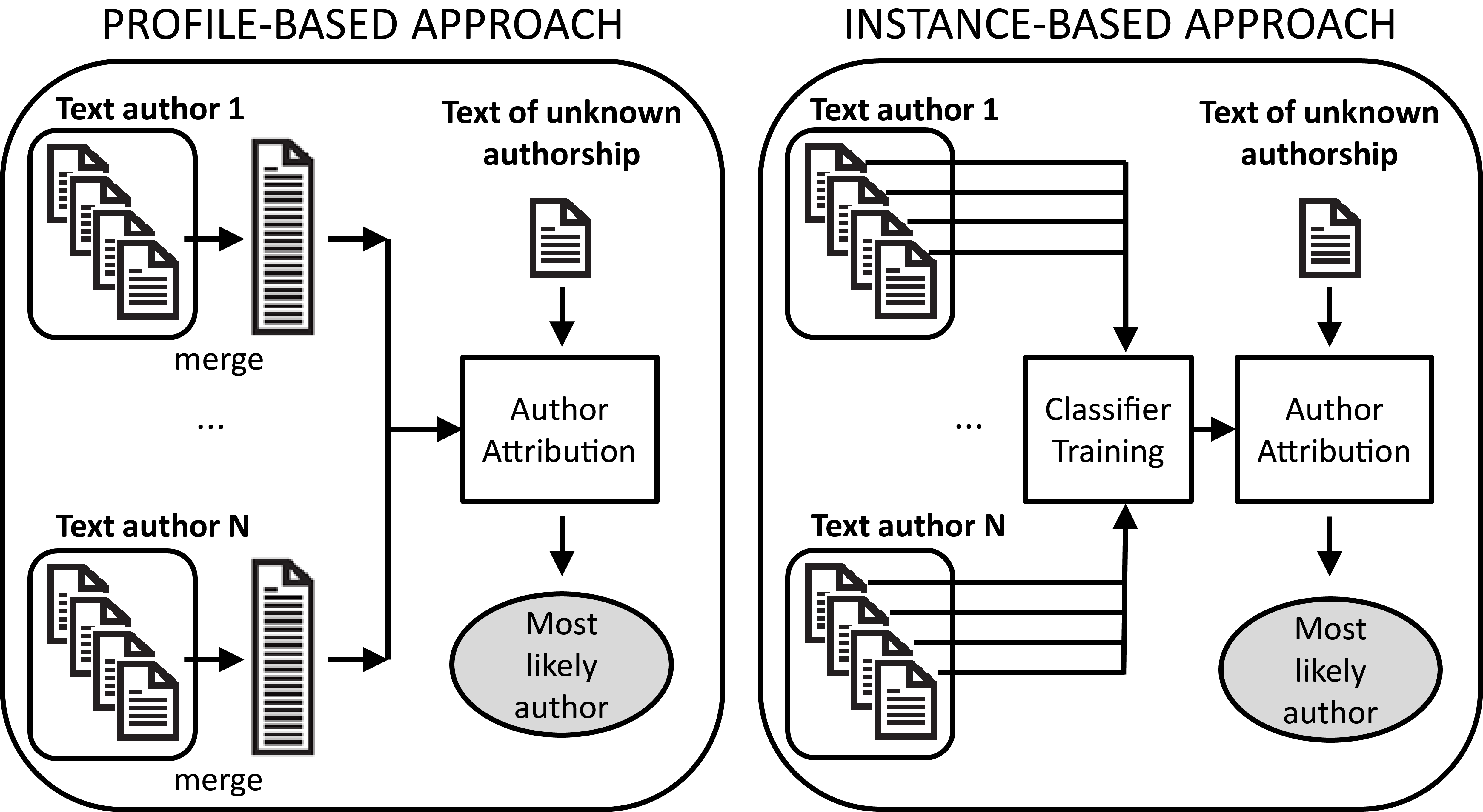}
\caption{Approaches to handle the set of documents available per author.}
\label{fig:approaches_models}
\end{figure}

Table \ref{tab:SOA} provides a summary of existing studies in online authorship identification.
Studies differ in the features used and the type of classifiers employed (Figure~\ref{fig:SOA}).
Among the features employed, these include for example
stylometric features \cite{Stamatatos2009} or $n$-grams \cite{Peng03}.
Stylometric features capture the writing style at different levels: character and word level (lexical), sentence and paragraph level (structural), topic level (content), punctuation and function words level (syntactic), and unique elements of a particular author (idiosyncratic).
%
A $n$-gram represents a sequence of $n$ elements next to each other in a text. The elements can be of different nature, for example a sequence of characters, words, symbols, syllables, etc.
To handle the available texts per author, two possible approaches are employed (Figure~\ref{fig:approaches_models}):
profile-based and instance-based \cite{Bouanani2014}.
In \textit{profile-based} approaches, all texts from an author are concatenated into a single text. This single text is used to extract the properties of the author’s style. In such way, the representation of an author could include text instances of different nature like formal texts and informal texts, creating a more comprehensive profile. Also, this approach can handle the problem of data imbalance and/or lack of enough data.
%
%
In the literature, this approach is implemented by using probabilistic techniques such as Naive Bayes, or distance measures.
In \textit{instance-based} approaches, on the other hand, every individual text is analysed separately, and a group of features from each particular text instance is extracted. The sets of features of every text instance are then used to train a model that represents the identity of the author.
This requires multiple training text samples per author in order to develop an accurate model, which may not be always available in forensic scenarios.
Instance-based approaches are usually implemented using clustering algorithms or machine learning classifiers such as Support Vector Machines (SVMs), Artificial Neural Networks (ANNs), or Decision Trees (DTs).

Some studies have been conducted specifically using short digital texts
such as tweets \cite{Okuno14} or SMS \cite{Donais13,Ragel13}.
The unique study concerned with tweet analysis \cite{Okuno14} reported an accuracy of 53.2\% in identifying the authorship among 10000 users.
A modified version of $n$-grams was used to handle short texts, where $n$-grams were weighted based on the length of the elements.
The amount of tweets per user was 120, which is in the same range than our database. On the other hand, the number of users is obviously much higher, which may be the reason of the lower accuracy in comparison to the present paper.
The work \cite{Donais13} reported an accuracy of 20.25\% in the classification of 2000 messages taken from an SMS corpus with 81 authors, with maximum 50 messages per author. 
Finally, the work \cite{Ragel13} employed a SMS corpus from 70 persons, with at least 50 SMS per person (some persons had several hundreds of messages).
As features, they employed unigram counts, which is simply the frequency of each word available in the text string.
Following the profile-based method of evaluation,
they carried out a number of different experiments varying the number of users in the database, and the size of training and test data.
For example, an experiment with 20 users having more than 500 SMS each for training gave an accuracy of about 88\% when the testing set contained 20 SMS per user. If only one SMS is available for testing, accuracy goes down to about 41\%.
They also studied the effect of reducing the size of the training set. With 100 training messages per author, accuracy barely goes above 61\% when 20 SMS per user are stacked for testing. With 50 training messages, accuracy goes below 48\%.
The authors acknowledge that these results may be positively biased, since they deliberately chose the SMSs having the maximum length.
All in all, these results show that authorship identification using text with a limited number of characters still remains a challenge.

\begin{figure}[htb]
\centering
        \includegraphics[width=0.4\textwidth]{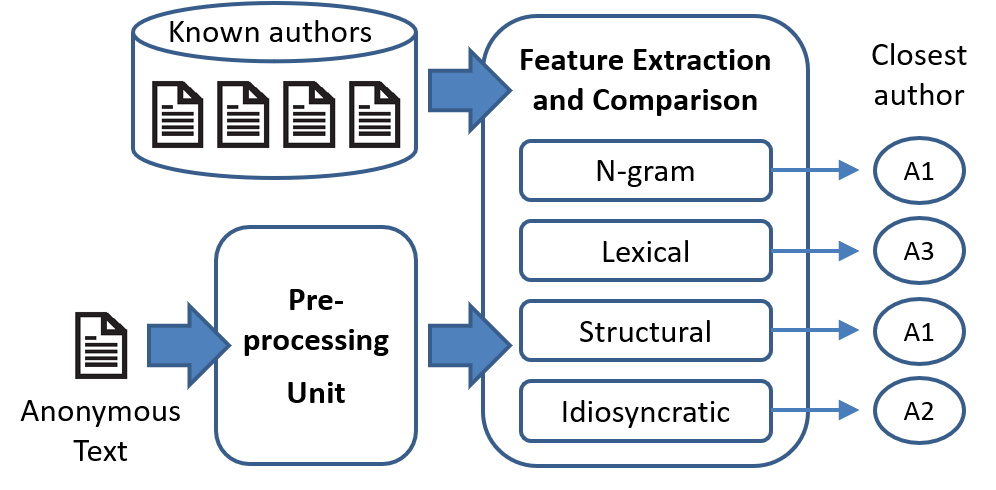}
\caption{Structure of the author identification system.}
\label{fig:system}
\end{figure}

\section{Methods for Writer Identification}

Here, we employ the most popular features for authorship attribution: $n$-grams and stylometric features.
In Figure~\ref{fig:system}, the overall model of the author identification system is depicted.

\subsection{$N$-Grams}

A $n$-gram represents a sequence of $n$ elements next to each other in a text. The elements can be of any nature, for instance a sequence of characters, words, symbols, syllables etc.
For example, the sentence ``the pen is blue'', when analyzed at the word level with $n$=2 would become a vector of $n$-grams with the form: [(the,pen),(pen,is),(is,blue)].
Next, the frequency of each $n$-gram element is calculated in the whole text, and the values are then used to represent the document as a vector.
%
%
The popular choice of this feature is explained by its scalability and language independence. It has been chosen for studies in different languages apart from English, e.g. \cite{Ragel13,Okuno14}.

Several studies have tried to establish what value should be assigned to $n$ to successfully capture the style of an author
\cite{Houvardas06,Wright17}, with experiments showing that
an increment in accuracy is observed as $n$ is augmented, but after 5 the accuracy tends not to improve significantly more.
In the present paper, both character $n$-grams and word $n$-grams will be tested, with $n$ between 2 and 4.
$N$-grams have been chosen as they can cope with the length of a tweet, misspellings, differences in language, as well as the presence of other symbols such as emojis or punctuation (given the fact that $n$-gram are not restricted to just letters).
%

\subsection{Stylometric Features}

Stylometry 
investigates features related to the distribution of words, the use of punctuation, the grammar, the structure of the sentence or paragraph and so on. Typically, the set of features to be analysed in a text are divided into five categories \cite{Stamatatos2009}: lexical, structural, content-specific, syntactic and idiosyncratic.
\textit{Lexical} features describe the set of characters and words that an individual uses. Such features include for example the distribution of uppercase characters, special characters, the average length of words used, the average of words used per sentence as well as other characteristics. This set describes the vocabulary richness of an author. 
\textit{Structural} features inform about the way the writer organizes the elements in a text, such as the number of paragraphs and sentences, or their average length. Here we also include indicators of whether the author includes greetings and farewell in an email corpus, for example.
\textit{Content-specific} refer to the frequency of particular keywords in the text.
This category is particularly handy when it comes to a corpus extracted from forums or topic-specific sources.
Despite being extremely insightful when it comes to content-monitoring in relation e.g. to terrorism and cyber-paedophilia, in a more general context such as Twitter posts, the features have been proved useless as they depend on a topic and environment \cite{Stamatatos2009}.
\textit{Syntactic} features are concerned with the syntax of the text, such as punctuation and function words. Function words are the words which help defining the relationships between the elements of a sentence. For this reason, they are also the most common words found in any text.
Unfortunately, given the length of a tweet, such features do not contribute
significantly to the representation of such texts either.
Finally, \textit{idiosyncratic} features aim at capturing
elements which are unique to the author. Such features include for instance the set of misspelt words, abbreviations, use of emojis or other special characters.

\begin{figure}[htb]
\centering
        \includegraphics[width=0.4\textwidth]{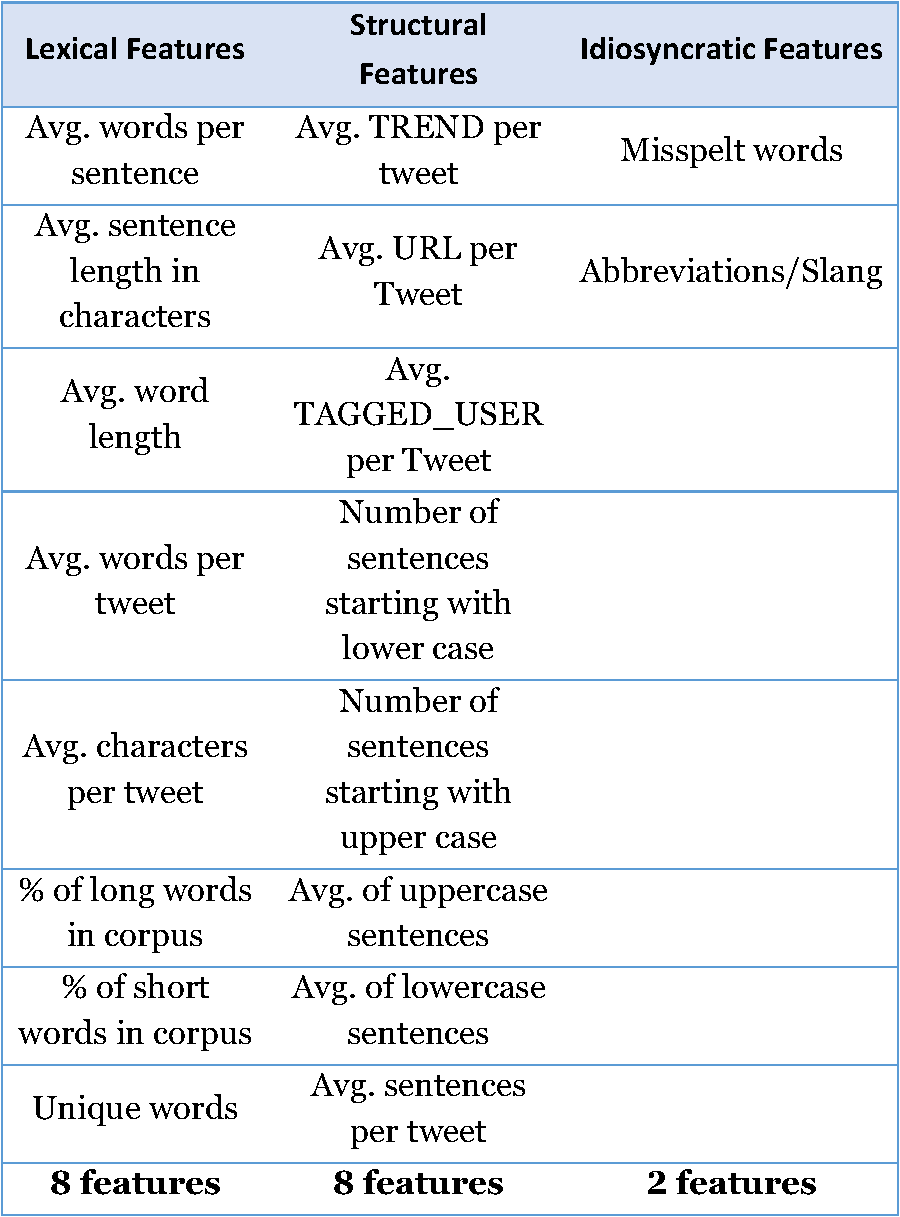}
\caption{Stylometric features employed.}
\label{fig:stylometric_features}
\end{figure}

The particular stylometric features chosen for this work are given in Figure~\ref{fig:stylometric_features}.
As indicated above,
%
content-specific features have not been taken into consideration as the set of tweets in our database are random, thus they do not follow a specific topic.
Also, given the length of a tweet, syntactic features are not reliable to use due to the lack of enough information to establish an author´s profile.
A number of features not typically present in the literature have been also added, since they are particularly useful to represent a tweet, e.g. average presence of 
URLs or tagged users.

\begin{figure*}[htb]
\centering
        \includegraphics[width=0.9\textwidth]{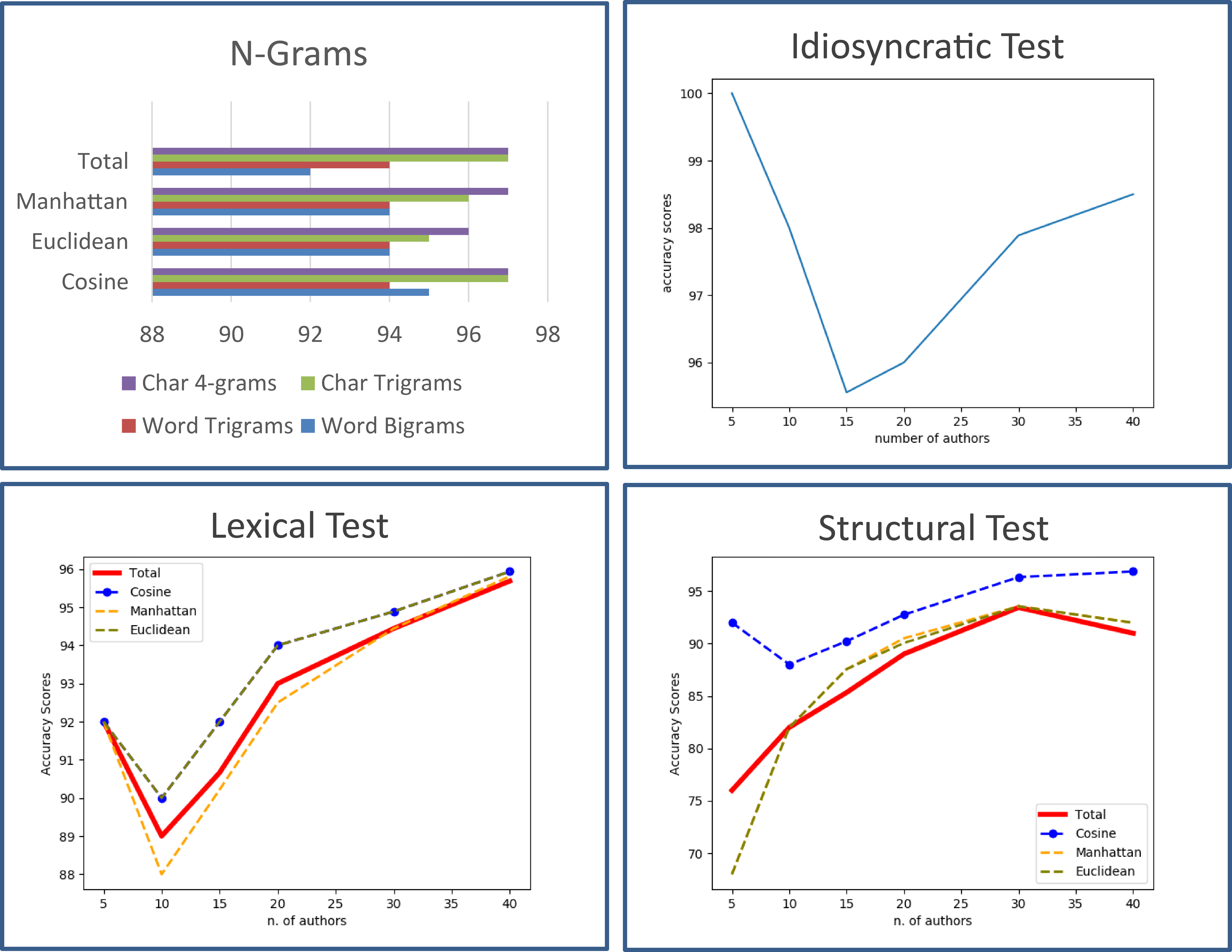}
\caption{Results of writer identification. Best seen in color.}
\label{fig:results}
\end{figure*}

\section{Database and Experimental Protocol}

Given the unavailability of a ready-to-use datasets of generic tweets, a dataset has been built from scratch. Twitter allows to download public tweets through their API.
The elements of the Twitter API have been accessed through Tweepy, an open source Python library 
which allows to download real-time tweets along with metadata, such as date of creation and data about the users, as long as they are public. 
A Python script has been written in order to 
retrieve tweets which meet the conditions “language == (English)” and “tweet.isRetweet == false”.
The program ran for 5 days, producing a list of 
40 users and 120-200 tweets per user.
%
%
%
The collected data has been split into 2 sets: 70\% of the text of each author is used to construct the corpus of known authorship, while the remaining
30\% is used as test data (of unknown authorship).

Raw tweets are pre-processed to remove elements such as user tags, hashtags and URLs, since
these elements can compromise the accuracy of some of the chosen methods.
This is because 
trending tags, hashtags and URLs are likely to be used by many users, hence they do not contribute to successfully identify a particular writer.
Pre-processed tweets are then analysed according to the different features employed.
A \textit{profile-based} approach (Figure~\ref{fig:approaches_models}) has been used in this work. The available tweets of a user are combined and treated as a whole corpus, where the features will be extracted from.
This approach has been chosed due to the shortness of tweet texts. Even though several features will be extracted, the length of the texts could be limiting the performance. A previous study on short pieces of text (SMS) \cite{Ragel13} has demonstrated that a higher accuracy is achieved if such type of messages are joint into one longer corpus.
Stylometric features are extracted with the Natural Language Toolkit (NTLK) library \cite{Bird09}.
Lexical and structural features are represented as numerical vectors, and the distance between vectors of different authors is used as metric of similarity.
In this work, three distance measures are evaluated: Cosine, Euclidean and Manhattan.
Idiosyncratic features, on the other hand, aim at capturing unique flaws/characteristics in the writing style of an authors (in our case, frequently misspelt words or slang words).
Here, features are represented as a vocabulary or set of words particular of each author.
To compare sets of different authors, the intersection between them is computed. The larger the intersection, the more similar are the two sets.
In addition to stylometric features, $n$-grams are extracted as well, as the literature shows their efficiency regardless of the length of the texts and context.
In this work, character and word $n$-grams are used, with $n$ between 2 and 4. All $n$-grams from a text are concatenated into a single vector, and the similarity between vectors of different authors is computed with distance measures: Cosine, Euclidean and Manhattan.

\section{Results}

%
In order to assess the reliability of the methods, the accuracy of each feature is assessed individually. The accuracy evaluates the number of correct writer predictions over the total number of predictions.
The tests have been run for a series of iterations, where the number of authors has been incremented gradually from 5 to 40 in steps of 5.
Results are given in Figure~\ref{fig:results}. When distance measures are employed, results are shown for each distance, and for the average (total) of all three distances.

In the experiments with $n$-grams, character $n$-grams with $n$ = 3, 4 and word $n$-grams with $n$ = 2, 3 have been used. Experiments with $n$-grams have been carried out with a subset of 10 users only.
From Figure~\ref{fig:results}, it can be seen that the best results are obtained with character $n$-grams, with no significant difference between $n$=3 or 4. Accuracy in both cases is 97\% with the Cosine distance. Word $n$-grams give a slightly worse accuracy of 94-95\%.
In Table~\ref{tab:results_ngrams},
the shortest distances recorded with each author are reported.
Numbers in red represent trials where the unknown author is incorrectly identified (the author mistakenly identified as the unknown author is given in brackets).
Numbers in blue indicate that the unknown author has correctly been identified, but another author of the database has the same distance value.
It can be noticed that a certain subset of authors seems to be mistakenly identified more often than others, e.g. authors 3 and 4.

With lexical features, it can be observed that the best results are obtained with the Cosine and the Euclidean distance. The Manhattan distance shows slightly worse results.
Lexical features have an accuracy between 92-96\% depending on the number of users, except a drop with 10 users. This seems an anomaly that could be due to a data problem with the set of authors selected. Overall, the test shows a high accuracy. Nonetheless, the anomaly in the results should be further investigated.
Such anomaly is also observed with the structural features, but only with the Cosine distance. Here, the Cosine distance is superior to the other two distances, with an accuracy in the same range than the lexical features.
Regarding idiosyncratic features, we observed that the property ``misspelt words'' is contributing to the results to a much more extent than ``abbreviations/slang''. This is because no abbreviations or slang were detected for the majority of the author in the database.
This may be the reason of the ``down-up'' behaviour observed in the idiosyncratic plot. %
Also, the majority of the authors holding idiosyncratic features are located in the second half of the dataset (the author selection is not random, but in increasing order, from author 1 to 40).
The accuracy reached for 40 users with idiosyncratic features is higher than with lexical or structural features (98.5\%).

\begin{table*}[htb]
\begin{center}
\begin{tabular}{c}

\includegraphics[width=0.9\textwidth]{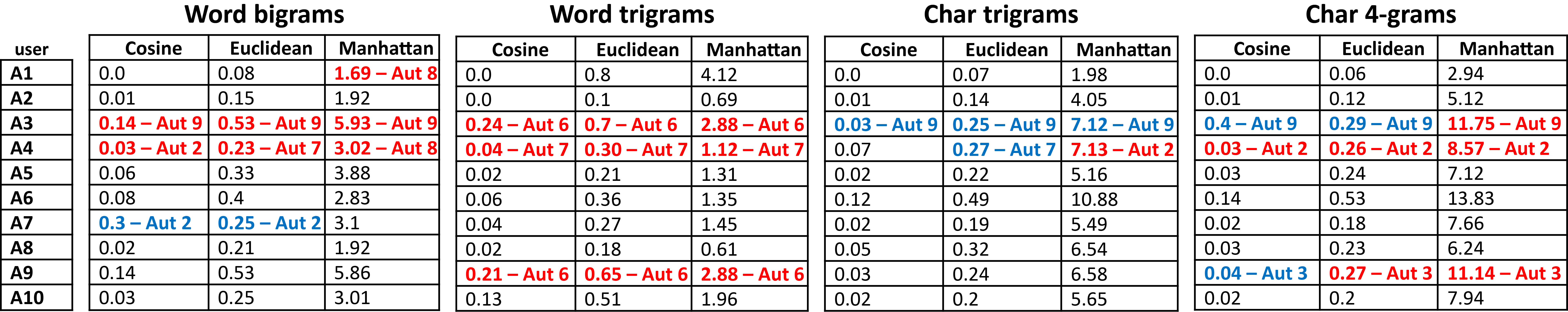} \\

\end{tabular}

\end{center}
\caption{Detail of results of writer identification with $n$-grams. Best seen in color.}
\label{tab:results_ngrams}
\end{table*}

\section{Discussion}

Authorship attribution has its roots in the 19th century, with 
many studies carried out even before the technology revolution. In the early days, the exploration was purely an application of stylometric techniques \cite{Rong2006}, with the main objective being to identify the author of literary works \cite{Williams75} or essays \cite{HOLMES95}.
Due to the increase of text availability in digital forms such as short messages, blogs, posts, etc.,
%
the subject of authorship analysis has evolved, becoming
an open question in many other fields including cybercrime, law enforcement, education, fraud detection and so on.
Because of the rapid development of cybercrimes, it is thus of huge interest the development of methods for authorship attribution to aid in forensic investigations \cite{Nirkhi2013}.

Accordingly, this work carried out an study of authorship attribution using texts in digital form.
Rarely the analyst have at disposal long texts per suspect, especially when the source of information is popular social media platforms such as Twitter or Facebook. Some platforms even limit the number of characters per message to a few hundreds only.
Here, we concentrate on short texts (Twitter posts), which are limited to 280 characters.
Different machine experts based on the two most popular features for authorship attribution are employed ($n$-grams and stylometric features).
They capture properties of the writing style at different levels, such as sequences of individual elements ($n$-grams), information about the characters and words that an individual uses (lexical), the way the writer organizes elements such as sentences and paragraphs (structural),
or particular elements which are unique of each author,
such as slang or misspelt words (idiosyncratic).

Experimental results are given using a self-acquired database of 40 users, with 120-200 tweets per user. Accuracy of the features are assessed individually, with the features capturing particularities of each author (idiosyncratic) showing slightly superior classification performance (98.5\%), followed by $n$-grams (97\%).
In any case, the worst-performing features have an accuracy higher than 92\%, showing the potential of this field of study.
The shortness of such texts may imply a huge obstacle in authorship attribution.
On the other hand, such ways of communication entail
alterations to the formal writing rules, which could hold positive connotations for our purpose. For example, abbreviations, slang word or special symbols can provide evidence of idiolectic nature \cite{Stamatatos2009}. In our experiments, indeed, such idiolectic/idiosyncratic features are the ones providing the highest accuracy.

Despite these performance numbers, several research questions remain open.
As a starting point, the model should be evaluated on a larger set of authors to further test the correlation between the efficiency of the methods and the authors set size \cite{Ragel13,Okuno14}.
The issue of some specific authors being consistently mistaken by others should also be looked at.
Feature fusion could be a path to counteract this effect, and to enhance the performance when larger sets are to be used \cite{[Fierrez18]}.
Other enhancement would be the use of methods to handle different lengths in the text.
Imbalance in the available amount of text per writer has not been addressed in this work.
The number of tweets per user ranges between 120 and 200. We suspect that this may be the reason of the inferior performance of lexical and structural features, which are dependant on the length of the texts. The best performing features in our experiments (idiosyncratic and $n$-grams) do not rely on the length of the text.

\section*{Acknowledgment}

Author F. A.-F. thanks the Swedish Research Council for
funding his research.
Author N. M. has been funded by European Social Fund via IT Academy programme.
Authors acknowledge the CAISR program
of the Swedish Knowledge
Foundation.

\bibliographystyle{IEEEtran}



\end{document}